\documentclass{ngsm2024} 

\usepackage{amsmath,amsfonts,bm}

\def\eqref#1{equation~\ref{#1}}

\def\1{\bm{1}}

\def\vx{{\bm{x}}}

\def\mU{{\bm{U}}}
\def\mV{{\bm{V}}}
\def\mW{{\bm{W}}}

\DeclareMathAlphabet{\mathsfit}{\encodingdefault}{\sfdefault}{m}{sl}
\SetMathAlphabet{\mathsfit}{bold}{\encodingdefault}{\sfdefault}{bx}{n}

\usepackage[utf8]{inputenc} 
\usepackage[T1]{fontenc}    
\usepackage{url}            
\usepackage{enumitem}
\usepackage{wrapfig}
\usepackage{booktabs}       
\usepackage{amsfonts}       
\usepackage{nicefrac}       
\usepackage{microtype}      
\usepackage{xcolor}         
\usepackage{adjustbox}
\usepackage{multirow}
\usepackage{textcomp}
\usepackage{pifont}
\usepackage[font=small, tableposition=top]{caption}
\usepackage[misc]{ifsym}
\usepackage{floatrow}
\usepackage{array}
\usepackage{arydshln}
\usepackage{verbatim}
\usepackage{comment}
\usepackage[subtle]{savetrees}
\newcolumntype{H}{>{\setbox0=\hbox\bgroup}c<{\egroup}@{}}
\newcommand{\lowrank}{Low-Rank}

\def\equationautorefname~#1\null{Eq.~(#1)\null}

\title[]{Investigating Low-Rank Training in Transformer Language Models: Efficiency and Scaling Analysis}

\optauthor{
    \Name{Xiuying Wei}  \Email{xiuying.wei@epfl.ch} \\
    \addr CLAIRE, EPFL
    \AND
     \Name{Skander Moalla} \Email{skander.moalla@epfl.ch} \\
    \addr CLAIRE, EPFL
    \AND
    \Name{Razvan Pascanu}
    \Email{razp@google.com} \\
    \addr Google DeepMind
    \AND
    \Name{Caglar Gulcehre}
    \Email{caglar.gulcehre@epfl.ch} \\
    \addr CLAIRE, EPFL
    }

\begin{document}

\maketitle

\begin{abstract}%
State-of-the-art LLMs often rely on scale with high computational costs, which has sparked a research agenda to reduce parameter counts and costs without significantly impacting performance.
Our study focuses on Transformer-based LLMs, specifically applying low-rank parametrization to the computationally intensive feedforward networks (FFNs), which are less studied than attention blocks. In contrast to previous works, (i) we explore low-rank parametrization at scale, up to 1.3B parameters; (ii) within Transformer language models rather than convolutional architectures; and (iii) starting from training from scratch.
Experiments on the large RefinedWeb dataset show that low-rank parametrization is both efficient (e.g., 2.6$\times$ FFN speed-up with 32\% parameters) and effective during training. Interestingly, these structured FFNs exhibit steeper scaling curves than the original models. Motivated by this finding, we develop the wide and structured networks surpassing the current medium-sized and large-sized Transformer in perplexity and throughput performance. Our code is available at \url{https://github.com/CLAIRE-Labo/StructuredFFN/tree/main}.
\end{abstract}


\section{Introduction}
Transformer language models~\citep{transformer} have gained significant attention for their performance and scalability. These models have grown from hundreds of millions of parameters~\citep{gpt-2} to hundreds of billions~\citep{gpt-3, llama-2, megatron}, increasing the need for efficient training and inference.  While much research focuses on attention, \emph{feed forward network}s (FFNs) account for over 60\% of the model's parameters and FLOPs, significantly impacting latency.
Low-rank parametrization, as one of the very popular structured matrices, is an important technique to make linear layer efficient. However, they have not yet been thoroughly explored at sufficient scales as a modification in modern LLM architectures.

In this work, we investigate low-rank matrices for FFN blocks from initialization on recent Transformer language models ranging from 110M to 1.3B parameters. Specifically, by using low-rank parametrization with 32\% of the parameters of FFN, the training speed of the 1.3B model can be boosted by 1.35$\times$ with only a 1 PPL increase. Interestingly, the low-rank parametrization has steeper loss scaling curves than the traditional Transformer at its optimal trade-off~\autoref{fig:performance}(a), suggesting a high potential for even better performance at larger scales. Finally, combined with \citet{gqa} for attention, we design wide and structured networks with slightly better PPL and maximum throughput performance under the same training FLOPs (e.g., 8\% and 17\% throughput boost on medium- and large-sized models). We hope our findings and results shed new light on the study of efficient NLP architectures.

\vspace{-0.5em}
\section{Related work}

Low-rank matrices have been widely used to decompose pre-trained weights for downstream compression~\cite{laser} and to construct adapters for efficient fine-tuning~\cite{lora, lora_theory} like LoRA. LoRA uses a low-rank approximation to reduce trainable parameters, while \citet{laser} selectively applies low-rank decomposition to well-trained weights.

Several works investigate low-rank training. \citet{implicitrank} argues that dense layers naturally converge to low-rank solutions during training, making this parametrization ideal. Early works like \citet{denil2013predicting, lowrankcnn} showed high efficiency of low-rank training. Some studies \citet{lowrankortho, lowranktrained, lowranknonlinearinit} adapt rank during training and suggest regularizers for better accuracy. \citet{lowrankfd} propose spectral initialization and aligned weight decay for matrix products. \citet{lowranknonlinearinit} suggest learning the initialization of low-rank matrices with data. However, these studies mainly focus on ResNets~\cite{resnet} rather than recent LLMs.

In this paper, we train low-rank matrices with a fixed rank as a replacement for the FFN linear layers of recent Transformers from scratch and investigate the performance of the new architecture. Formally, the low-rank parametrization of a linear layer can be given as $\mW\vx \approx \mU(\mV\vx)$, where $\mW$ is the original weight, $\vx$ is the input, $\mU \in \mathbb{R}^{M \times R}$, $\mV \in \mathbb{R}^{R \times N}$, and $R < \min(M, N)$. This reduces parameter count and FLOPs from $M \cdot N$ to $(M+N) \cdot R$.

\vspace{-0.5em}
\section{Experiments}\label{sec:experiments}

\subsection{Settings}
\paragraph{Implementation}We replace only the FFN modules with low-rank parametrization, as the attention module is well-studied~\cite{gqa, mqa}. We use ranks that are half or a quarter of the original hidden state dimension, reducing FFN parameters to 63\% or 32\% of the original size. The first FFN module remains unchanged to avoid significant performance degradation. For initialization, we follow the spectral initialization suggested by prior works~\cite{lowrankfd}. 

\paragraph{Training} We use a basic Transformer architecture~\cite{attention, gpt-2} with Rotary Embedding~\cite{roformer} and a basic FFN module composed of two linear layers and a GeLU activation function. Our model ranges from 110M to 1.3B parameters and is trained on the RefinedWeb dataset~\cite{refinedweb}. We randomly select 0.5B tokens as validation set while the number of training tokens is allocated based on the scaling law~\cite{scalinglaw}. We measure training FLOPs as in Megatron~\cite{megatronlm}, including all matrix multiplications. Hyperparameters, such as learning rates and global batch size, are set according to recent studies~\cite{mamba, opt}. Details are summarized in \autoref{tab:baseline_config}.\begin{table}[htbp]
\caption{Model and Training configuration. We report the number of layers~(\textbf{\#Layer}), hidden states dimension~(\textbf{Width}), training tokens~(\textbf{Tokens)}), global batch size in number of tokens~(\textbf{Batch}), peaking learning rate~(\textbf{LR}), and total training steps~(\textbf{Steps}).}
\centering
\begin{adjustbox}{max width=0.9\textwidth}

\small
\begin{tabular}{llllllHlHlHHl}

\toprule
\textbf{Name} & \textbf{Size} & \textbf{Width} & \textbf{Layers} & \textbf{Tokens} & \textbf{Batch} & \textbf{Mini Batch Size} & \textbf{LR} & \textbf{Global Batch Size Time} & \textbf{Steps} & \textbf{Time on GPU} & \textbf{GPUs} & \textbf{Training FLOPs}\\ \midrule
Transformer-s  & 110M & 768 & 12 & 2.2B & 0.5M & 64/64  & 6.0e-4 & 3.11s  & 4.2K  & 1 GPU $\sim$5h   & 2 (3h)  & 1.69e+18 \\ \midrule
Transformer-m & 335M & 1024 & 24 & 6.7B  & 0.5M & 32/64  & 3.0e-4   & 11s    & 13K & 1 GPU $\sim$45h  & 4 (11h) & 1.55e+19 \\ \midrule
Transformer-l  & 729M & 1536 & 24 & 14.6B  & 0.5M & 16/32 & 2.5e-4   & 19.65s & 28K & 1 GPU $\sim$165h & 8 (20h) & 7.03e+19\\ \midrule
Transformer-xl & 1274M & 2048 & 24 & 25.5B & 0.5M & 16/32 & 2.0e-4   & 29.46s for bs & 49K & 1 GPU $\sim$400h & 8 (50h) & 2.10e+20\\ \midrule
\end{tabular}
\end{adjustbox}
\label{tab:baseline_config}
\end{table}

\subsection{Efficiency and accuracy performance}
We evaluate both the efficiency and accuracy performance of low-rank parametrization in FFN. First, as shown in \autoref{fig:performance}(b), with increasing FFN width, GPU resources can be utilized more thoroughly, and this parametrization can bring a 1.4$\times$ and 2.6$\times$ speed-up with 63\% and 32\% of the parameters, respectively, compared to the width of 1536. 

Second, in \autoref{tab:complete_efficient_linear_layer}, we observe that this parametrization results in about a 0.4 PPL increase on Transformer-xl with a 15\% reduction in training time, and about a 1.0 higher PPL with a 1.35$\times$ speed-up for the whole model.
\begin{table}[htbp]
\footnotesize
\caption{Performance of low-rank parametrization with 63\% and 32\% of the original FFN module's parameters, where $R$ indicates the rank. Note that the total structured FFN is not exactly 63\% of the original because we don't replace the first FFN module.}
\centering
\begin{adjustbox}{max width=0.8\linewidth}
\begin{tabular}{llHHHHlHHcccHc}
\toprule
\multirow{2}{*}{\textbf{Architecture}} & \bf{Model} & FLOPs (T) & fwd+bwd  & prefill & decoding  & \bf{FFN} & FLOPs (T) & prefill (ms) & \multicolumn{3}{c}{\bf {Training}} & \multirow{2}{*}{\bf{Loss}} & \multirow{2}{*}{\bf PPL}\\
\cmidrule{9-12}
 & \bf{Size (M)} & FLOPs (T) & fwd+bwd  & prefill & decoding  & \bf{Size (M)} & FLOPs (T) & prefill (ms) & \bf {Tokens (B)}  & \bf {FLOPs} & \bf {Time (h)} &  & \\
\midrule
    \bf Transformer-s & 110 & 134.62 & - & - & - & 
    57 & 59.37 & 1.40 & 
    2.2 &  1.69e+18 & 4.0 & 
    3.2569 &  25.97 \\
    \midrule
    \lowrank{} (R=384) & 90 & 114.21 & - & - & - & 
    37 & 38.96 &  & 
    2.2 & 1.44e+18 & 3.8 & 
    3.3017  & \bf 27.16\\
    \midrule
    \lowrank{} (R=192) & 74 & 97.20 & - & - & - & 
    21 & 21.96 &  & 
    2.2 & 1.22e+18 & 3.6 & 
    3.3748 & \bf 29.22\\

    \midrule
    \bf Transformer-m & 335 & 403.80 &  & - & - & 
    201 & 211.11 & & 
    6.7 & 1.55e+19 & 32.5 &
    2.9062 & 18.29 \\
    \midrule
    \lowrank{} (R=512) & 263 & 327.93 & - & - & - & 
    129 & 135.24 &  & 
    6.7 & 1.26e+19 & 29.6 &
    2.9508 & \bf 19.12 \\
    \midrule
    \lowrank{} (R=256) & 202 &264.71  & - & - & - & 
    69 & 72.02 & & 
    6.7 & 1.01e+19 & 26.9 &
    3.0251 & \bf 20.60 \\
    \midrule
    \bf Transformer-l & 729 & 843.19 &  & - & - & 
    453 & 474.99 & & 
    14.6 & 7.03e+19 & 130.5 &
    2.6594 & 14.29 \\
    \midrule
     \lowrank{} (R=768) &566  & 672.49 & - & - & - & 
    290 & 304.29 &  & 
    14.6 & 5.61e+19 & 113.6 &
    2.6957 & \bf 14.82 \\ 
    \midrule 
    \lowrank{} (R=384) & 431  & 530.24 & - & - & - & 
    155 & 162.04  &  & 
    14.6 & 4.42e+19 & 100.0 &
    2.7527 & \bf 15.69 \\
    \midrule
    \bf Transformer-xl & 1274 & 1440.91 & - & - & - & 
    805 & 844.42 &6.11 & 
    25.5 & 2.10e+20 & 352.2 &
    2.5226 & 12.46\\
    \midrule
    \lowrank{} (R=1024) & 985 & 1137.44 & - & - & - & 
    516 & 540.96 & - & 
    25.5 & 1.66e+20 & 302.2 &
    2.5541 & \bf 12.86 \\
    \midrule
    \lowrank{} (R=512) & 744 &884.56  & - & - & - & 
    275 & 288.07 & - & 
    25.5 & 1.29e+20 & 260.2 &
    2.6062 & \bf 13.55 \\

\bottomrule
\end{tabular}
\end{adjustbox}
\label{tab:complete_efficient_linear_layer}
\end{table}

\vspace{-1.5em}
\subsection{Scaling analysis}
\begin{figure}[t]
    \centering
    \includegraphics[width=0.9\textwidth]{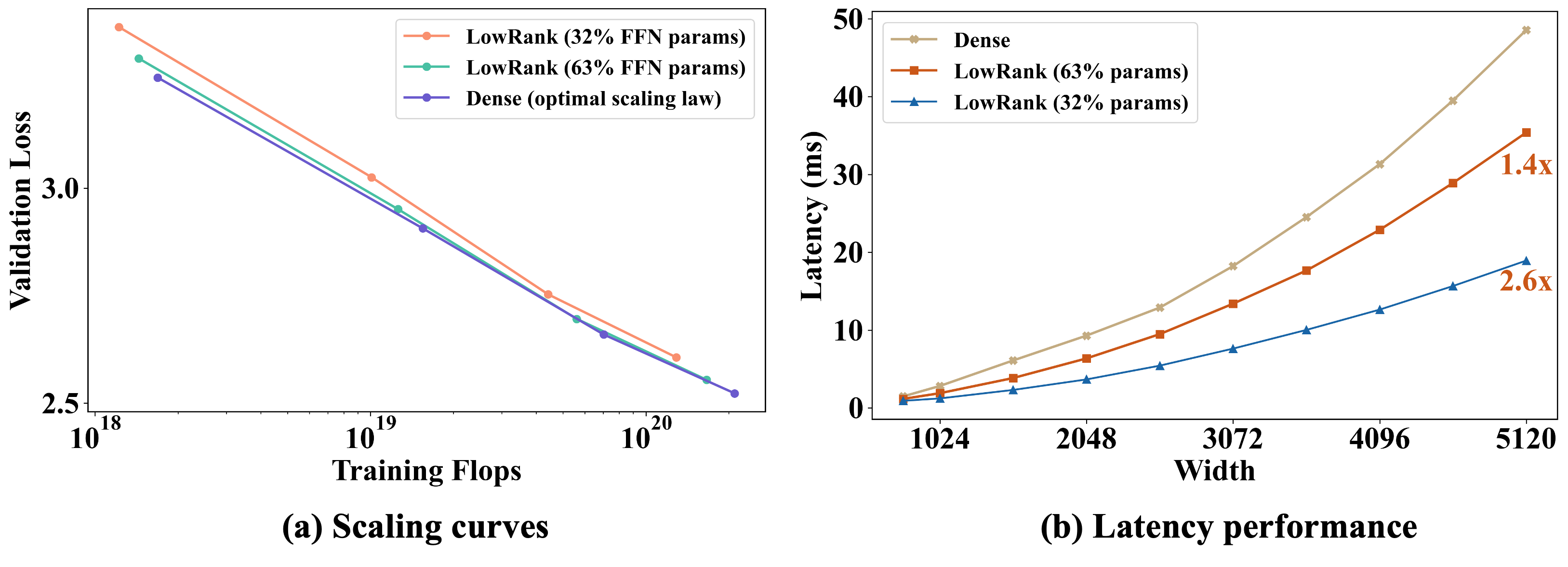}
    \caption{(a): The training scaling curves between the standard Transformer and the modified version with low-rank parametrization, which retains 63\% and 32\% of the original parameters, respectively. (b): FFN latency performance across different widths, measured on 30,000 tokens.}
    \label{fig:performance}
    \vspace{-1em}
\end{figure}

From \autoref{fig:performance}(a), it can be seen that the low-rank parametrization gets closer to the baseline when the model size increases. Technically, we observe that: 
(i) \emph{The low-rank parametrization exhibits steeper scaling curves compared to the dense networks, indicating significant potential for these efficient designs in LLMs.} (ii) \emph{ The scaling curve of 32\% parameters of FFN is steeper than the 63\% parameters of FFN highlights the scaling potential of highly structured large models.} (iii) \emph{Given fixed training FLOPs budget, a wider and structured network with more tokens may achieve comparable or superior performance to dense networks at the optimal trade-off.} 

The scaling curves can be further optimized: (1) they are not drawn at their optimal training-compute trade-off unlike the baseline. (2) Only the FFN is made structured, while attention remains dense, contributing more to the model's performance. The second point also explains why the current 32\% parameter curve shows a larger validation loss than the 63\% parameter curve under the same training FLOPs. This motivates us to further reduce attention using existing techniques in \autoref{sec: wide_sparse}.

\subsection{Wide and Structured network}
\label{sec: wide_sparse}
Motivated by the scaling curves, we reduce both the attention and FFN and create a wide and structured network, as shown in \autoref{tab:new_arch}. This approach aims to enhance efficiency with a much smaller network, achieving an 8\% and 17\% maximum throughput boost compared to medium- and large-sized GQA~\cite{gqa} models while maintaining or slightly improving perplexity.

\begin{table}[htbp]
    \centering
    \caption{
    We compare the performance of GQA and our wide, structured networks. \textbf{Left}: TP indicates the maximum throughput measured for a generation length of 256. \textbf{Right}: Dimensions of various components, including hidden states, FFN intermediate states, attention, and KVCache. GQA's intermediate size is increased to match parameters, as in \citet{llama-3}.}
    \begin{adjustbox}{max width=\textwidth}
    \begin{tabular}{llllll:llll}
    \toprule
    \textbf{Method} & \textbf{\#Param} & \textbf{Training FLOPs} & \textbf{Tokens} & \textbf{PPL} & \textbf{TP (256)} & \textbf{Hidden} & \textbf{Intermediate} & \textbf{Attention} & \textbf{KV} \\ 
    \midrule
     Transformer-m & 335M & 1.55e+19 & 6.7B & 18.29 & 30229 & 1024 & 4096 & 1024 & 1024 \\
     Transformer-m (GQA) & 335M & 1.55e+19 & 6.7B &  18.23 & 84202 & 1024 & 4864 & 1024 & 256 \\
     Low-Rank (R=512) & 219M & 1.55e+19 & 10.6B  & \textbf{17.89} & \textbf{91147} & 1024 & 4864 & 512 & 256 \\
     \midrule
     Transformer-l & 729M & 7.03e+19 & 14.6B & 14.29 & 23351 & 1536 & 6144 & 1536 & 1536 \\
     Transformer-l (GQA) & 729M & 7.03e+19 & 14.6B &  14.40 & 64737 & 1536 &7424 & 1536 & 256 \\
     Low-Rank (R=768) & 464M & 7.03e+19 & 22.3B &  \textbf{14.27} & \textbf{75930} & 1536 & 7424 & 768 & 256 \\
    \bottomrule
    \end{tabular}
    \end{adjustbox}
    \label{tab:new_arch}
\end{table}

\vspace{-1.5em}
\section{Conclusion and Limitation}
\label{sec:conclusions}
In this paper, we investigate low-rank parametrization in the FFN of Transformer language models. Training such structured models from scratch shows promising scaling curves and efficiency. However, we have not explored its optimal scaling laws and have only limited our exploration to the language aspect. Studying the upper limits and other applications of low-rank training would also be very valuable.



\bibliography{sample}


\end{document}